\documentclass{article}
\usepackage{spconf,amsmath,epsfig, amssymb, blkarray, subfig, wrapfig, algorithmic}
\usepackage{enumitem}
\usepackage{color}
\newcommand{\remi}[2]{{\color{blue}#2}}
\title{HANDLING UNCERTAINTIES IN SVM CLASSIFICATION}
\name{\'Emilie NIAF$^{1,2}$, R\'emi FLAMARY$^{3}$, Carole LARTIZIEN$^{2}$, St\'ephane CANU$^{3}$}
\address{$^{1}$INSERM U556, Lyon, 69424, France\\
	    $^{2}$CREATIS, UMR CNRS 5220; INSERM U1044; INSA-Lyon; UCBL, Villeurbanne, 69621, France\\
        $^{3}$LITIS EA 4108, INSA-Universite de Rouen, Saint-Etienne-du-Rouvray, 76801, France}

\begin{document}
%\ninept
%
\maketitle
\begin{abstract}
This paper addresses the pattern classification problem arising when available target data include some uncertainty information.
Target data considered here is either qualitative (a class label) or quantitative (an estimation of the posterior probability).
Our main contribution is a SVM inspired formulation of this problem
allowing to take into account class label through a hinge loss
  as well as probability estimates using $\varepsilon$-insensitive cost function together with a minimum norm (maximum margin) objective. This formulation shows a dual form leading to a quadratic problem and allows the use of a representer theorem and associated kernel.
 The solution provided can be used for both decision and posterior probability estimation.
 Based on empirical evidence our method outperforms regular SVM in terms of  probability predictions and classification performances. 
%Support Vector Machines (SVM) are commonly used methods for supervised classification. Traditionally, the learning step is performed over a fully annotated set of points, and possible labels uncertainties are not considered.
%We propose to introduce these uncertainties in the learning step by allowing labels to be the probability values to belong to one class. We thus present a new formulation of the SVM classification problem leading to a new optimization problem. Our approach allows not to discard uncertain training samples as well as to moderate the influence of samples with high uncertainties.
%Example classification problems with uncertain data are considered. Results demonstrate that this method significantly outperforms regular SVM in terms of probability predictions and classification performances.
\end{abstract}
\section{Introduction}
\label{sec:intro}

In the mainstream supervised classification scheme, an expert is
required for labelling a set of data used then as inputs for training the classifier. However, even for an
expert, this labeling task is likely to be difficult in many
applications. In the end the training data set may contain inaccurate classes for some examples, which
leads to non robust classifiers\cite{stempfel2009learning}. For instance, this is often the case in medical imaging where radiologists have to outline what they think are malignant tissues over medical images without access to the reference histopatologic information. We propose to deal with these uncertainties by introducing probabilistic labels in the learning stage so as to: 1. stick to the real life annotation problem, 2. avoid discarding uncertain data, 3. balance the influence of uncertain data in the classification process. \\
Our study focuses on the widely used Support Vector Machines (SVM)
two-class classification problem \cite{schölkopf2002learning}. This
method aims a finding the separating hyperplane maximizing the margin between the examples
  of both classes. 
Several mappings from SVM scores to class membership
probabilities have been proposed in the literature
\cite{plattprobabilistic, Sollich00probabilisticmethods}. In our approach, we
  propose to use both labels and probabilities as input thus learning
  simultaneously a classifier and a probabilistic output. Note that
  the output of our classifier may be transformed to probability estimations
  without using any mapping algorithm. \\
In section~\ref{sec:psvm-formulation} we define our new SVM problem formulation (referred to as P-SVM) to deal with certain and probabilistic labels simultaneously. Section~\ref{sec:dual_form} describes the whole framework of P-SVM and presents the associated quadratic problem. Finally, in section~\ref{sec:examples} we compare its performances to the classical SVM formulation (C-SVM) over different data sets to demonstrate its potential. 

\section{Problem formulation}
\label{sec:psvm-formulation}
We present below a new
  formulation for the two-class classification problem dealing with
uncertain labels. Let $X$ be a feature space. We define $(x_i,
l_i)_{i=1\dots m}$ the learning dataset of input vectors
$(x_i)_{i=1\dots m}\in X$ along with their corresponding labels $(l_i)_{i=1\dots m}$, the latter of which being
\vspace*{-0.2cm}
\begin{itemize}
\item class labels: $l_i$ = $y_i \in \{-1,+1\}$ for  $i=1\dots n$ (in classification),
\vspace*{-0.2cm}
\item real values: $l_i$ = $p_i \in [0,1]$ for $i=n+1 \dots m$ (in regression).
\end{itemize}
\vspace*{-0.2cm}
$p_i$, associated to point $x_i$ allows to consider uncertainties
about point $x_i$'s class. We define it as the posterior probability for class 1.
\vspace*{-0.15cm}
\begin{center}
$p_i = p(x_i) = \mathbb{P}(Y_i=1\mid X_i=x_i)$. 
\end{center}
\vspace*{-0.15cm}
We define the associated pattern recognition problem as
\vspace*{-0.3cm} 
\begin{eqnarray}
\vspace*{-0.2cm} 
\label{eq:svm_etendu_primal1}
\displaystyle\min_w& \quad\frac{1}{2} \|w\|^2\\
%\displaystyle\min_{w} \frac{1}{2} \|w\|^2
%\end{equation}
%\vspace*{-0.2cm} 
\text{subject to}&
%\begin{equation}
\nonumber %\mbox{avec} 
\begin{cases}
 y_i(w^\top x_i+b) \geq 1, &i=1...n  \\
 z_i^{-} \leq  w^\top x_i+b \leq z_i^{+}, &i = n+1...m \vspace*{-0.2cm}
\end{cases}
\end{eqnarray}
Where boundaries $z_i^{-}$, $z_i^{+}$ directly depend on $p_i$.
This formulation consists in minimizing the complexity of the model while forcing good classification and
good probability estimation (close to $p_i$).
%The first constraint forces good classifications while the second forces probability predictions to remains close to the true probability.
Obviously, if $n=m$, we are brought back to the classical SVM problem formulation.

Following the idea of soft margin introduced in regular SVM to deal with the case of inseparable data, we introduce slack variables $\xi_i$. This measure the degree of misclassification of the datum $x_i$ thus relaxing hard constraints of the initial optimization problem which becomes
%To deal with the case of inseparable classification problems, we follow the soft-margin approach used for standard SVM (called C-SVM) and introduce slack variables, $\xi_i$, measuring the degree of misclassification of the datum $x_i$.
%The optimization  problem then becomes
\vspace*{-0.3cm} 
\begin{equation}
\vspace*{-0.2cm} 
\label{eq:svm_etendu_primal}
\displaystyle\min_{w,\xi, \xi^{-}, \xi^{+}} \frac{1}{2} \|w\|^2 + C\displaystyle\sum\limits_{i=1}^{n}\xi_i + \tilde{C}\displaystyle\sum\limits_{i=n+1}^{m}(\xi^{-}_i+\xi^{+}_i)
\end{equation}
\vspace*{-0.2cm} 
subject to
\vspace*{-0.1cm}
\begin{equation}
\nonumber %\mbox{avec} 
\begin{cases}
 y_i(w^\top x_i+b) \geq 1 - \xi_i, &i=1...n  \\
 z_i^{-} - \xi^{-}_i  \leq  w^\top x_i+b \leq z_i^{+} + \xi^{+}_i , &i = n+1...m\\
 0 \leq \xi_i, &i=1...n \\
 0 \leq \xi_i^{-} \mbox{ and } 0 \leq \xi_i^{+}, &i=n+1...m \\
\end{cases}
\end{equation} 
Parameters $C$ and $\tilde{C}$ are predefined positive real numbers controlling the relative weighting  of classification and regression performances.\\ %\vspace*{-0.1cm}
%the goal of making the margin large and, respectively, minimizing classification errors minimizing probability prediction errors.  controls the 
%\section{SVM extended to the probabilistic labels case}
%\label{sec:psvm-prob_explanation}
%\subsection{Problem formulation in the primal}
%Ideally, we try to find optimal parameters w and b such that, for i=n+1$\dots$ m:
% \begin{equation}
%\label{eq:pi}
%\begin{array}{l}
%p_i = \frac{1}{1 + e^{-a(w^\top  x_i+b)}} \\
%\end{array}
%\end{equation}
%Note that:
%\begin{center}
%$ \mathbb{P}(Y_i=0\mid X_i=x_i) = 1 - p_i = \frac{1}{1 + e^{a(w^\top  x_i+b)}}$\\
%\end{center} 
Let $\varepsilon$ be the labelling precision and $\delta$ the confidence we have in the labelling. Let's define $\eta$ = $\varepsilon$  + $\delta$.
%Then, a more realistic problem consists in finding optimal parameters w and b such that:
Then, the regression problem consists in finding optimal parameters $w$ and $b$ such that
\vspace*{-0.5cm} 
\begin{center}
$\mid \frac{1}{\displaystyle 1 + e^{-a(w^\top  x_i+b)}}  - p_i \mid <  \eta$ , \\
\end{center}
Thus constraining the probability prediction for point $x_i$ to remain around to $\frac{1}{1 + e^{-a(w^\top  x_i+b)}}$ within distance $\eta$ \cite{rupingsimple, grandvalet2006probabilistic, ruepingsvm}. 
The boundaries (where $w^\top  x_i+b = \pm 1$), define parameter $a$ as:
\vspace*{-0.3cm}
\begin{center}
   $a = \ln(\frac{1}{\eta} - 1)$
\end{center}
\vspace*{-0.2cm}
Finally:
\vspace*{0.1cm} \hspace*{-1cm} \resizebox{.5\textwidth}{!}{
\begin{tabular}{*{5}{c}}
  $\max(0, p_i- \eta)$ & $\leq$ & $\frac{1}{\displaystyle 1 + e^{-a(w^\top  x_i+b)}}$ & $<$ & $\min(p_i + \eta,1)$,	\\
  $\Longleftrightarrow$ $z_i^{-}$ & $\leq$ & $w^\top  x_i+b$ & $<$ & $z_i^{+}$, \\
\end{tabular} } \vspace*{0.3cm} 
%\begin{equation*}
%    \begin{array}{llll}
%     &\mbox{max}(0, p_i- \eta) &\leq \frac{1}{\displaystyle 1 + e^{-a(w^\top  x_i+b)}} &< \mbox{min}(p_i + \eta,1) ,\\
%     \Longleftrightarrow&  z_i^{-} &\leq w^\top  x_i+b &< z_i^{+}, \\
%    \end{array}
%\end{equation*} 
 
\hspace*{-0.6cm} 
where $ z_i^{-}= -\frac{1}{a}\ln(\frac{1}{p_i - \eta} -1)$ and $z_i^{+} = -\frac{1}{a}\ln(\frac{1}{p_i + \eta} - 1)$.
\section{Dual formulation}
\label{sec:dual_form}
%\subsection{Dual formulation}
We can \remi{then}{} rewrite the problem in its dual form, introducing Lagrange multipliers. We are looking for a stationary point for the Lagrange function $L$ defined as

$\textit{L}(w, b, \xi, \xi^{-}, \xi^{+}, \alpha, \beta, \mu^+, \mu^-, \gamma^{+}, \gamma^{-}) = $
\vspace*{-0.2cm}
\begin{equation*}
\label{eq:Lag}
\begin{array}{ll}
&\frac{1}{2} \|w\|^2 + C\displaystyle\sum\limits_{i=1}^{n}\xi_i + \tilde{C}\displaystyle\sum\limits_{i=n+1}^{m}(\xi^{-}_i+\xi^{+}_i)\\ 
&-\displaystyle\sum\limits_{i=1}^{n}\alpha_i(y_i(w^\top x_i+b) - (1 - \xi_i)) - \displaystyle\sum\limits_{i=1}^{n}\beta_i \xi_i\\
& - \displaystyle\sum\limits_{i=n+1}^{m}\mu_i^{-}((w^\top x_i+b) - (z_i^{-} - \xi_i^{-})) - \displaystyle\sum\limits_{i=n+1}^{m}\gamma_i^{-} \xi_i^{-} \\
&- \displaystyle\sum\limits_{i=n+1}^{m}\mu_i^{+}((z_i^{+} + \xi_i^{+}) - (w^\top x_i+b)) - \displaystyle\sum\limits_{i=n+1}^{m}\gamma^{+}_i \xi_i^{+} \\
\end{array}
\end{equation*}
with $\alpha \geq 0$, $\beta \geq 0$, $\mu^{+} \geq 0$, $\mu^{-} \geq 0$,$\gamma^{+} \geq 0$ and $\gamma^{-} \geq 0$\\
Computing the derivatives of $L$ with respect to $w, b, \xi, \xi^{-}$ and $\xi^{+}$ leads to the following optimality conditions:
%\begin{equation*}
%  \left\{
%    \begin{array}{rll}
%      C e_{1} & = &\alpha + \beta   \\
%      \tilde{C} e_{2}  &=& \mu^{-} + \gamma^{-} = \mu^{+} + \gamma^{+} \\
%      w &=&  \displaystyle\sum\limits_{i=1}^{n}\alpha_i y_i x_i - \displaystyle\sum\limits_{i=n+1}^{m}(\mu_i^{+} - \mu_i^{-})x_i \\
%      0 &=& \displaystyle\sum\limits_{i=n+1}^{m}(\mu_i^{+} - \mu_i^{-}) - \displaystyle\sum\limits_{i=1}^{n}\alpha_i y_i
%    \end{array}
%  \right.
%\end{equation*}
%Since $\beta \geq 0$, $\gamma^{+} \geq 0$, $\gamma^{-} \geq 0$, these conditions become: 
\begin{equation*}
  \left\{
    \begin{array}{rll}
      0 &\leq \alpha_{i} \leq C, &i=1...n \\
      0 &\leq \mu^{+}_{i} \leq \tilde{C}, &i=n+1...m \\
      0 &\leq \mu^{-}_{i} \leq \tilde{C}, &i=n+1...m \\
      w &=  \displaystyle\sum\limits_{i=1}^{n}\alpha_i y_i x_i - \displaystyle\sum\limits_{i=n+1}^{m}(\mu_i^{+} - \mu_i^{-})x_i &\\
      y^\top \alpha &= \sum\limits_{i=n+1}^{m}(\mu_i^+ - \mu_i^-)&
    \end{array}
  \right.
\end{equation*}
where $e_1 = [\underbrace{1\dots 1}_{\mbox{\footnotesize{n times }}} \underbrace{0\dots 0}_{\mbox{\footnotesize{(m-n) times }}}]^\top $ and $e_2 = [\underbrace{0\dots 0}_{\mbox{\footnotesize{n times }}} \underbrace{1\dots 1}_{\mbox{\footnotesize{(m-n) times }}}]^\top $.\\
Calculations simplifications then lead to\\
$\textit{L}(w, b, \xi, \xi^{-}, \xi^{+}, \alpha, \beta, \mu, \gamma^{+}, \gamma^{-})=$
\vspace*{-0.2cm}
\begin{eqnarray*}
 & -\frac{1}{2} w^\top  w +\displaystyle\sum\limits_{i=1}^{n}\alpha_i
+ \displaystyle\sum\limits_{i=n+1}^{m}\mu_i^{-}z_i^{-}
- \displaystyle\sum\limits_{i=n+1}^{m}\mu_i^{+}z_i^{+}\nonumber
\end{eqnarray*}
Finally, let $\Gamma$ = $[\alpha_1 \dots \alpha_n$ $\mu^{+}_{n+1} \dots \mu^{+}_m$ $\mu^{-}_{n+1} \dots \mu^{-}_m ]^\top $ be a vector of dimension $2m - n$. Then
\begin{center}
$w^\top w$ = $\Gamma^\top $ G $\Gamma$
\end{center}
\vspace*{-0.5cm}
where\\
\vspace*{-0.5cm}
\begin{center}
G = 
$
\left(
\begin{array}{lllcrr}
 &K_1 & -&K_2 & &K_2\\
 -&K_2^\top  & &K_3 & -&K_3 \\
 &K_2^\top  & -&K_3 & &K_3  
\end{array}
\right)
$
\end{center}
with\\
\vspace*{-0.4cm}
\begin{equation*}
\begin{array}{lll}
K_1 &=& (y_iy_jx_i^\top  x_j)_{i,j = 1\dots n},\\
K_2 &=& (x_i^\top  x_jy_i)_{i = 1\dots n, j = n+1\dots m},\\
K_3 &=& (x_i^\top  x_j)_{i,j = n+1\dots m},\\
\end{array}
\end{equation*}
The dual formulation becomes\\
\begin{equation}
\label{eq:svm_etendu_dual}
  \left\{
%  \begin{aligned}
    \begin{array}{lll}
      \displaystyle\min_{\Gamma}& \frac{1}{2} \Gamma^\top  G \Gamma - \tilde{e}^\top  \Gamma,\\ 
      &f^\top \Gamma=0  \\
      \mbox{ with }& \tilde{e} = [\underbrace{1 \dots 1}_{\mbox{n times}} \underbrace{-z_{n+1}^{+} \dots -z_{m}^{+}}_{\mbox{n-m times}}\underbrace{z_{n+1}^{-} \dots z_{m}^{-}}_{\mbox{n-m times}}]\\    
      \mbox{ with } & f^\top =[y^\top ,  \underbrace{-1 \dots -1}_{\mbox{n-m times}},\underbrace{1 \dots 1}_{\mbox{n-m times}}] \\
      \mbox{ and } &0 \leq \Gamma \leq [\underbrace{C \dots C}_{\mbox{n times}} \underbrace{\tilde{C} \dots \tilde{C}}_{\mbox{n-m times}}\underbrace{\tilde{C} \dots \tilde{C}}_{\mbox{n-m times}}]^\top 
    \end{array}
%    \end{aligned}
  \right.
\end{equation}
\section{Kernelization}
%\subsection{Kernelization}
%Formulation generalized to kernel functions
Formulations (\ref{eq:svm_etendu_primal}) and
(\ref{eq:svm_etendu_dual}) can be easily generalized by introducing
kernel functions. Let $k$ be a positive kernel satisfying Mercer's
condition and H the associated Reproducing Kernel Hilbert Space (RKHS). Within this framework equation (\ref{eq:svm_etendu_primal}) becomes
\vspace*{-0.2cm}
\begin{equation}
\displaystyle\min_{f,b,\xi, \xi^{-}, \xi^{+}} \frac{1}{2} \|f\|^2_H + C\displaystyle\sum\limits_{i=1}^{n}\xi_i + \tilde{C}\displaystyle\sum\limits_{i=n+1}^{m}(\xi^{-}_i+\xi^{+}_i)\\
\end{equation}
subject to
\begin{equation}
\nonumber %\mbox{avec} 
\begin{cases}
	y_i(f(x_i)+b) \geq 1 - \xi_i, &i=1...n  \\
 	z_i^{-} - \xi^{-}_i  \leq  f(x_i)+b \leq z_i^{+} + \xi^{+}_i , &i = n+1...m\\
    0 \leq \xi_i, &i=1...n \\
    0 \leq \xi_i^{-} \mbox{ and } 0 \leq \xi_i^{+} &i=n+1...m \\
\end{cases}
\end{equation}
Formulation (\ref{eq:svm_etendu_dual}) remains identical, with\\
$
\begin{array}{lll}
K_1 &=& (y_iy_jk(x_i, x_j))_{i,j = 1\dots n},\\
K_2 &=& (k(x_i, x_j) y_i)_{i = 1\dots n, j = n+1\dots m},\\
K_3 &=& (k(x_i, x_j))_{i,j = n+1\dots m},\\
\end{array}
$
%where $k(\bullet,\bullet)$.
\section{Examples}
\label{sec:examples}

In order to experimentally evaluate the proposed method for handling uncertain labels in SVM classification, we have simulated different data sets described below. In these numerical examples, a RBF kernel $k(u,v) = e^{-\| u-v\|^2 / 2\sigma^2}$ is used and $C = \tilde{C} = 100$. We implemented our method using the SVM-KM Toolbox \cite{SVM-KMToolbox}.
We compare the classification performances and probabilistic predictions of the C-SVM and P-SVM approaches.
In the first case, probabilities are estimated by using Platt's scaling algorithm \cite{plattprobabilistic} while in the second case, probabilities are directly estimated via the formula defined in (\ref{sec:psvm-formulation}): $P(y=1|x) = \frac{1}{1 + e^{-a(w^\top  x+b)}}$.
Performances are evaluated by computing
\vspace*{-0.3cm}
\begin{center}
\begin{itemize}
\item Accuracy (Acc) \\
Proportion of well predicted examples in
the test set (for evaluating classification).
%\frac{\text{number of true positives}+\text{number of true negatives}}{\text{numbers of true positives}+\text{false positives} + \text{false negatives} + \text{true negatives}}. 
\vspace*{-0.2cm}
\item Kullback Leibler distance (KL) \\
$D_{KL}(P||Q)= \displaystyle\sum\limits_{i=1}^{n}P(y_i = 1|x_i)\log(\dfrac{P(y_i = 1|x_i)}{Q(y_i = 1|x_i)})$
for probability distributions P and Q (for evaluating probability estimation).\\
%\item Alignement error\\
%$1 - \displaystyle\sum\limits_{i=1}^{n}\dfrac{ P(y_i = 1|x_i)Q(y_i = 1|x_i)}{\sqrt{Q(y_i = 1|x_i)^2}\sqrt{P(y_i = 1|x_i)^2}}$
\end{itemize}
\end{center}
\subsection{Probability estimation}
\label{subsec:prob_est}

\begin{figure}[!h]
\centering
\includegraphics[width=0.7\columnwidth, keepaspectratio]{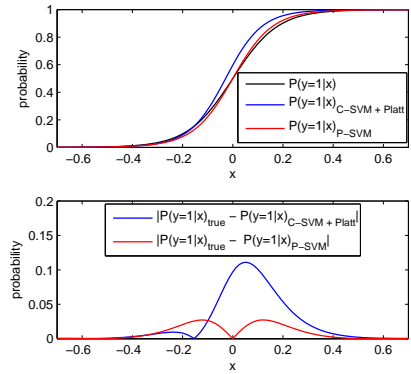}
%1D_not_normalized_prob_test_svm_9_2.eps
\vspace*{-0.3cm}
\caption{Probability estimations comparison. Top plot shows the true posterior probabilities with C-SVM and P-SVM estimations overlaying. Lower plot shows the distance between true probabilities and estimations.}
\label{fig:1D_prob_estim}
\end{figure}

We generate two unidimensional datasets, labelled '+1' and '-1', from normal distributions of variances  $\sigma_{-1}^2$= $\sigma_{1}^2$=0.3 and means $\mu_{-1}$=-0.5 and $\mu_{1}$=+0.5. Let's $(x^{l}_{i})_{i=1\dots n^{l}}$ denote the learning data set ($n^l$=200) and $(x^{t}_{i})_{i=1\dots n^{t}}$ the test set ($n^t$=1000). We compute, for each point $x_{i}$, its true probability $P(y_i=+1|x_i)$ to belong to class '+1'.
From here on, learning data are labelled in two ways, as follows
%\hspace*{-1cm}
\begin{enumerate}[label=\alph*), align=left, leftmargin=*, noitemsep]
\item For $i=1\dots n^l$, we get the regular SVM dataset by simply
  using a probability of 0.5 as the threshold for assigning class
  labels $y_i$ associated to point $x_i$. This is what would be done
  in practical cases when the data contains class membership
    probabilities and a SVM classifier is used.
\begin{equation}
  \label{eq:etiquetage}
  \begin{array}{llllll}
	\mbox{if } &P(y^{l}_{i} = 1|x^{l}_{i})&>& 0.5, &\mbox{then }& y^{l}_{i} = 1, \\
	\mbox{if } &P(y^{l}_{i} = 1|x^{l}_{i})&\leq& 0.5, &\mbox{then }& y^{l}_{i} = -1
  \end{array}
\end{equation}
This dataset $(x^l_{i},y^l_{i})_{i=1\dots n^l}$ is used to train the C-SVM classifier.
\item We define another data set $(x^l_{i},{\hat{y}}^l_{i})_{i=1\dots n^l}$ such that, for $i=1\dots n$,
\begin{equation}
  \label{eq:semi_etiq_prob}
  \begin{array}{lrlll}
	\mbox{if } &P(y^l_{i} = 1|x^l_{i})&>& 1 - \eta, &\mbox{then } {\hat{y}}^l_{i} = 1, \\ 
	\mbox{if } &P(y^l_{i} = 1|x^l_{i})&<& \eta, &\mbox{then } {\hat{y}}^l_{i} = -1,\\ 
	&{\hat{y}}^l_{i} &=& P(y^l_{i} = 1|x^l_{i}) &\mbox{otherwise}. 
 \end{array}
\end{equation}
If the probability values are sufficiently close to 0 or 1 (closeness being defined by the precision and confidence), we admit that they belong respectively to class -1 or 1. This probabilistic dataset $(x^l_{i},{\hat{y}}^l_{i})_{i=1\dots n^l}$ is used to train the P-SVM algorithm.\\ 
\end{enumerate}
\vspace*{-0.7cm}
We compare our two approaches using the test set $(x^t_{i})_{i=1\dots n^t}$.
As we know the true probabilities $(P(y^t_{i} = 1|x^t_{i}))_{i=1\dots n^t}$,
we can estimate the probability prediction error (KL).
Figure ~\ref{fig:1D_prob_estim} shows the probability predictions performances improvement shown by the P-SVM: the true probabilities (black) and P-SVM estimations (red) are quasi-superimposed (KL=0.2) whereas Platt's estimations are less accurate (KL=11.3). %5 et 20
\vspace*{-0.4cm}
\vspace*{-0.6cm}
\subsection{Noise robustness}
\label{subsec:prob_est}
\begin{figure}[!h]
\centering
\includegraphics[width=0.3\textwidth, keepaspectratio]{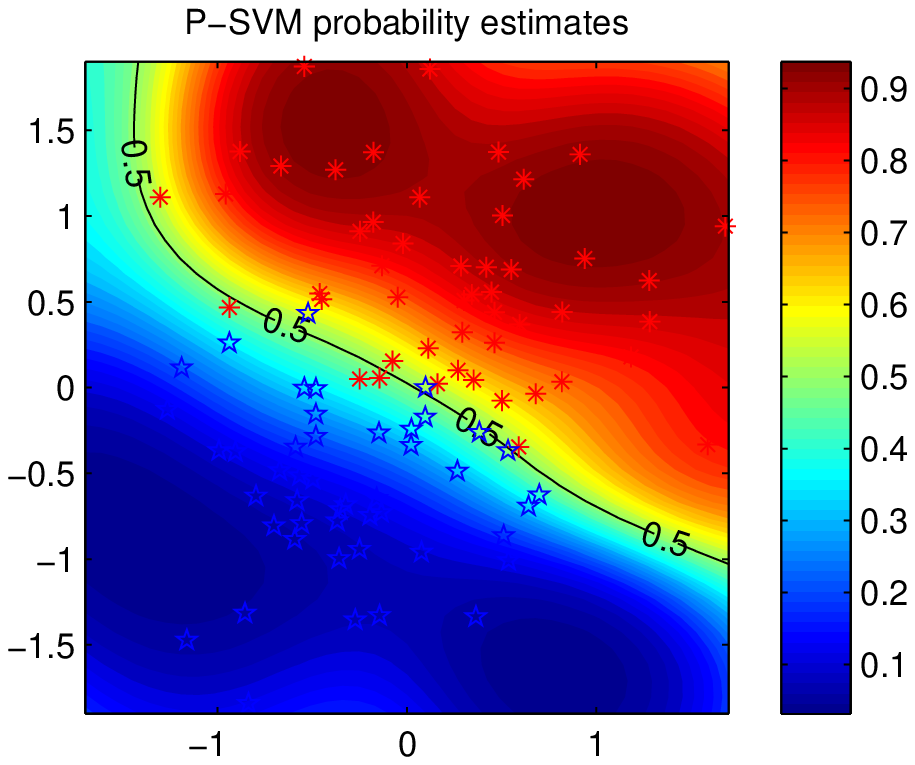}
\includegraphics[width=0.3\textwidth, keepaspectratio]{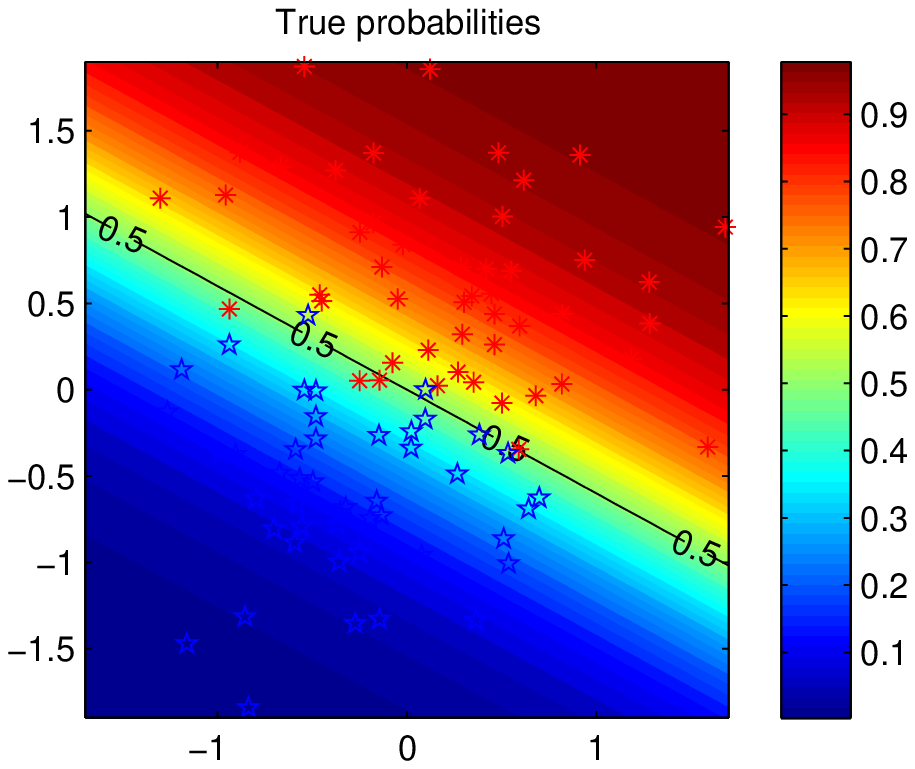}
%\end{figure}
%\begin{figure}[!h]
%\centering
\includegraphics[width=0.3\textwidth, keepaspectratio]{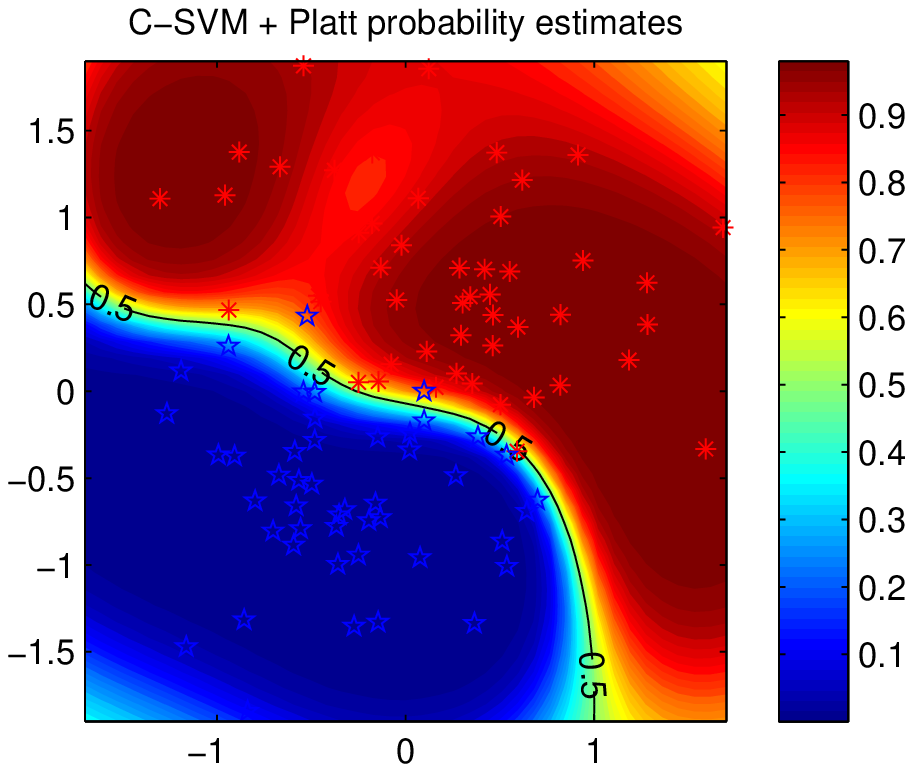}
\caption{Probability estimations of C-SVM and P-SVM over a grid using noisy learning data (uniform noise, amplitude 0.1). Noisy learning data are plotted in blue (class '-1') and red (class '1') stars.} 
% Performances obtained for 1000 test point are: Acc$_{\textrm{P-SVM}}$ = 99\% versus Acc$_{\textrm{C-SVM}}$ = 95\%,  % KL$_{\textrm{P-SVM}}$ = 3.6 versus KL$_{\textrm{C-SVM}}$ = 95}
\label{fig:2D_proba}
\end{figure}

We generate two 2D datasets, labelled '+1' and '-1', from normal
distributions of variances  $\sigma_{-1}^2$=$\sigma_{1}^2$=0.7 and
means $\mu_{-1}$ = (-0.3, -0.5) and $\mu_{1}$=(+0.3, +0.5). As in the
previous experiment, we compute class '1' membership probability for
each point $x^{l}$ of the learning data set. We simulate
classification error by artificially adding a centered uniform noise ($\delta$ of amplitude 0.1), to the probabilities, such that for $i=1\dots n$,
\vspace*{-0.15cm}
\begin{center}
$\hat{P}(y_i = 1|x_i) = P(y_i = 1|x_i) + \delta_i$. 
\end{center}
\vspace*{-0.15cm}
We then label learning data following the same scheme as described in (\ref{eq:etiquetage}) and (\ref{eq:semi_etiq_prob}). 
Figure~\ref{fig:2D_proba} shows the margin location and probabilities
estimations using the two methods over a grid of values. Far from
learning data points, both probability estimations are less accurate,
this being directly linked to the choice of a gaussian
kernel. However, P-SVM classification and probability estimations
obtained for 1000 test points, are clearly more alike the ground truth
(Acc$_{\textrm{P-SVM}}$ = 99\% , KL$_{\textrm{P-SVM}}$ = 3.6) than
C-SVM (Acc$_{\textrm{C-SVM}}$ = 95\%, KL$_{\textrm{C-SVM}}$ =
95). Contrary to P-SVM which, by combining both classification and
regression, predicts good probabilities, C-SVM is sensitive to
classification noise and is no more converging to the Bayes rule as
seen in~\cite{stempfel2009learning}.\vspace{-0.3cm}

\begin{figure}[!h]
\centering
\includegraphics[width=0.5\textwidth, keepaspectratio]{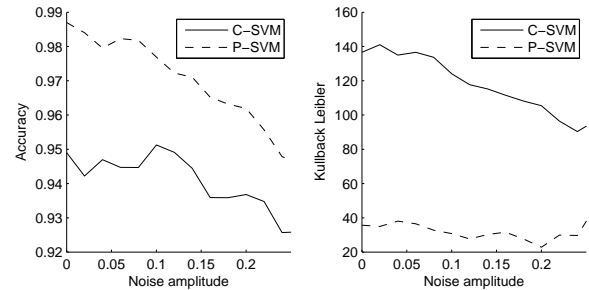}
\caption{Noise impact on P-SVM and C-SVM classification performances}
\label{fig:KL_Accuracy_versus_noise}
\end{figure}

Figure~\ref{fig:KL_Accuracy_versus_noise} shows the impact of noise amplitude on classifiers performances (values are averaged over 30 random simulations). Even if noise increases, classifications and probability predictions performances of the P-SVM remain significantly higher than those of C-SVM. 
\vspace{-.4cm}
\vspace*{1cm}
%\begin{figure}[htb]
%
%\begin{minipage}[b]{1.0\linewidth}
%  \centering
%  \centerline{\epsfig{figure=image1.eps,width=8.5cm}}
%%  \vspace{2.0cm}
%  \centerline{(a) Result 1}\medskip
%\end{minipage}
%%
%\begin{minipage}[b]{.48\linewidth}
%  \centering
%  \centerline{\epsfig{figure=image3.eps,width=4.0cm}}
%%  \vspace{1.5cm}
%  \centerline{(b) Results 3}\medskip
%\end{minipage}
%\hfill
%\begin{minipage}[b]{0.48\linewidth}
%  \centering
%  \centerline{\epsfig{figure=image4.eps,width=4.0cm}}
%%  \vspace{1.5cm}
%  \centerline{(c) Result 4}\medskip
%\end{minipage}
%%
%\caption{Example of placing a figure with experimental results.}
%\label{fig:res}
%%
%\end{figure}
%
%% To start a new column (but not a new page) and help balance the last-page
%% column length use \vfill\pagebreak.
%% -------------------------------------------------------------------------
%\vfill
%\pagebreak
%\vspace*{-0.5cm}
\section{CONCLUSION}
This paper has presented a new way to take into account both qualitative and quantitative target data
by shrewdly combining both SVM classification  and regression loss. 
Experimental results show that our formulation %is works well and  
can perform very well on simulated data for discrimination as well as posterior probability estimation. 
This approach will soon be applied on clinical data thus allowing to assess its usefulness in computer assisted diagnosis for prostate cancer. 
Note that this framework initially designed for probabilistic labels
can also be generalized to other dataset involving quantitative data
as it can be used for instance to estimate a conditional cumulative distribution function.
%http://www.unavarra.es/metma3/Papers/PDFS_ORAL/Pawlowsky.pdf
%\section{REFERENCES}
\label{sec:ref}

\small
\bibliographystyle{IEEEbib}
\bibliography{refs}

\end{document}